\documentclass{article}
\usepackage{amssymb}
\usepackage{hyperref}
\usepackage{multicol, multirow}
\usepackage{makecell}

\usepackage{times}
\usepackage{xcolor}
\usepackage{xspace}
\usepackage{acronym}
\usepackage{lipsum}
\usepackage{multicol}
\usepackage{graphicx}
\usepackage{caption}
\usepackage{amsmath}
\usepackage{cleveref}
\usepackage{amssymb}
\usepackage{bm}
\usepackage{booktabs}
\usepackage{tabularx}
\usepackage{comment}
\usepackage{soul}
\usepackage{wrapfig}

\crefname{algorithm}{Alg.}{Algs.}
\Crefname{algocf}{Algorithm}{Algorithms}
\crefname{section}{Sec.}{Secs.}
\Crefname{section}{Section}{Sections}
\crefname{table}{Tab.}{Tabs.}
\Crefname{table}{Table}{Tables}
\crefname{figure}{Fig.}{Figs.}
\Crefname{figure}{Figure}{Figures}
\crefname{equation}{Eq.}{Eqs.}
\Crefname{equation}{Equation}{Equations}
\crefname{appendix}{Appx.}{Appxs.}
\Crefname{appendix}{Appendix}{Appendices}

\makeatletter
\DeclareRobustCommand\onedot{\futurelet\@let@token\@onedot}
\def\@onedot{\ifx\@let@token.\else.\null\fi\xspace}
\def\eg{\emph{e.g}\onedot}

\def\etal{\emph{et al}\onedot}
\makeatother

\definecolor{phcolor}{RGB}{188, 188, 188}
\definecolor{tycolor}{RGB}{205, 92, 92}

\newcommand{\method}{ControlVLA\xspace}

\newcommand{\mbold}[1]{\boldsymbol{#1}}

\acrodef{vae}[VAE]{Variational Autoencoder}
\acrodef{ddpm}[DDPM]{Denoising Diffusion Probabilistic Model}
\acrodef{ddim}[DDIM]{Denoising Diffusion Implicit Model}
\acrodef{mlps}[MLPs]{Multi-layer Perceptrons}
\acrodef{mdp}[MDP]{Markov Decision Process}
\acrodef{cnn}[CNN]{Convolutional Neural Network}
\acrodef{vla}[VLA model]{Vision-Language-Action models}
\acrodef{rpn}[RPN]{Region Proposal Network}
\acrodef{vlm}[VLM]{Vision-Language Model}

\usepackage[preprint]{corl_2025} %

\title{\method: Few-shot Object-centric Adaptation for Pre-trained Vision-Language-Action Models}

\author{Puhao Li$^{1,2}$, Yingying Wu$^{1,2}$, Ziheng Xi$^1$, Wanlin Li$^2$, Yuzhe Huang$^2$, Zhiyuan Zhang$^2$, \\ 
\textbf{Yinghan Chen$^{2,3}$, Jianan Wang$^4$, Song-Chun Zhu$^{1,2,3}$, Tengyu Liu$^{2, \dagger}$, Siyuan Huang$^{2, \dagger}$} \\[4pt]
$^1$Tsinghua University \quad $^2$State Key Lab of General Artificial Intelligence, BIGAI \\
$^3$Peking University \quad $^4$Astribot Inc. \quad $^\dagger$Corresponding author\\[2pt]
{\hypersetup{urlcolor=orange}\url{https://controlvla.github.io}}\vspace{-10pt}
} %

\begin{document}
\maketitle

\begin{abstract}
Learning real-world robotic manipulation is challenging, particularly when limited demonstrations are available. Existing methods for few-shot manipulation often rely on simulation-augmented data or pre-built modules like grasping and pose estimation, which struggle with sim-to-real gaps and lack extensibility. While large-scale imitation pre-training shows promise, adapting these general-purpose policies to specific tasks in data-scarce settings remains unexplored. To achieve this, we propose \method, a novel framework that bridges pre-trained VLA models with object-centric representations via a ControlNet-style architecture for efficient fine-tuning. Specifically, to introduce object-centric conditions without overwriting prior knowledge, \method zero-initializes a set of projection layers, allowing them to gradually adapt the pre-trained manipulation policies.
In real-world experiments across 6 diverse tasks, including pouring cubes and folding clothes, our method achieves a 76.7\% success rate while requiring only 10-20 demonstrations --- a significant improvement over traditional approaches that require more than 100 demonstrations to achieve comparable success. 
Additional experiments highlight \method's extensibility to long-horizon tasks and robustness to unseen objects and backgrounds. 
\end{abstract}


\section{Introduction}
Robotic manipulation in the real world remains a fundamental challenge, particularly when learning skills from limited demonstrations. While recent advances in robotic manipulation~\citep{fu2024mobile, li2024ag2manip, liu2024rdt, zhu2023viola, chen2024g3flow, wang2024gendp, hsu2024spot, chi2024universal, ma2022vip, jiang2024dexmimicgen, brohan2023rt, yang2024learning, li2025maniptrans, huang2024embodied, li2023gendexgrasp, li2024grasp} have shown promise, current methods still demand extensive training data and struggle to efficiently adapt to new tasks and environments with few demonstrations. This limitation significantly hinders the deployment of robots in diverse real-world scenarios, where large amounts of task-specific training data are often impractical or prohibitively expensive.

To tackle this, previous works~\citep{torne2024reconciling, mandlekar2023mimicgen, jiang2024dexmimicgen, mu2024robotwin} augment expert demonstrations in simulation to enhance policy learning. However, these approaches typically assume \textit{a priori} knowledge of object and environment CAD models, as well as precise 3D pose estimations---requirements often impractical in real-world scenarios. Currently, imitation learning has achieved impressive success in manipulation~\citep{brohan2023rt, fu2024mobile, chi2024universal, chi2023diffusion, liu2024rdt, nair2023r3m, team2024octo, kim2024openvla},  primarily due to its scalability and capability to acquire skills across a wide range of scenarios without relying on external \textit{a priori} knowledge. 

In particular, \textbf{pre-trained general-purpose \ac{vla}}~\citep{brohan2022rt, brohan2023rt, kim2024openvla, team2024octo, liu2024rdt, o2023open} has emerged as a promising approach for enabling generalizable robot behaviors across various tasks and environments. However, fine-tuning these \ac{vla} for downstream tasks remains data-intensive, as substantial amounts of task- and environment-specific data are still required to adapt to the visual and action domains of the target task~\citep{black2410pi0, kim2024openvla, team2024octo, liu2024rdt}. On a separate front, \textbf{object-centric representations} has shown potential in improving data efficiency for learning expert policies~\citep{zhu2023viola, zhu2023learning, hsu2024spot}. By focusing on relevant object properties (\eg, shape, size, and position), object-centric representations reduce the complexity of the input observation space. This approach enhances policy robustness to changes in object pose and instance, while also making the policies less susceptible to real-world noise compared to pixel-level features. Despite this, existing methods still require hundreds of demonstrations to learn a task~\citep{zhu2023viola, zhu2023learning}, mainly due to the lack of an action prior from \ac{vla} in data-scarce scenarios.

\begin{figure}
    \centering
    \includegraphics[width=\linewidth]{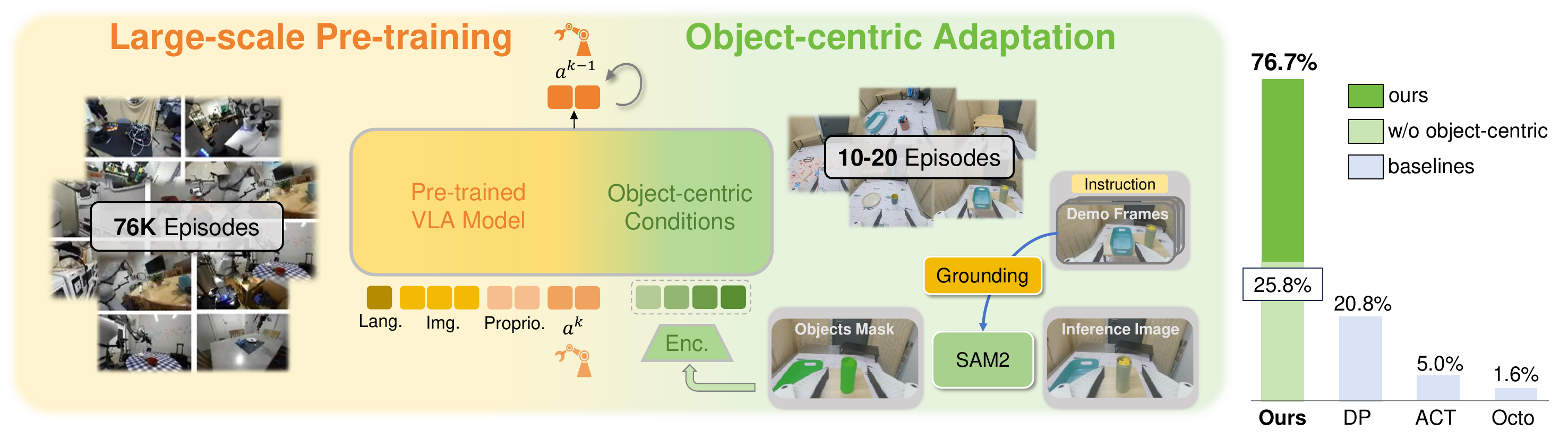}
    \caption{\textbf{\method bridges pre-trained manipulation policies with object-centric representations via ControlNet-style efficient fine-tuning.} \method requires only 10–20 demonstrations to achieve 76.7\% task success rate, significantly surpassing baseline's 20.8\% success rate.}
    \label{fig:teaser}
    \vspace{-16pt}
\end{figure}

To this end, we introduce \textbf{\method}, a novel learning framework that combines pre-trained \ac{vla} with object-centric representations for efficient few-shot learning. By integrating object-centric representations into a pre-trained \ac{vla}, our approach leverages both the rich prior knowledge from large-scale pre-training and the data efficiency of object-centric learning. Inspired by Zhang~\etal~\citep{zhang2023adding}, \method introduces \textbf{additional cross-attention layers with zero-initialized key-value (KV) projection weights}, allowing expert policies to acquire task-specific skills while progressively integrating object-centric representations. This design ensures that the policy focuses on task-relevant concepts without compromising the generalization or action quality of the pre-trained \ac{vla}. The zero-initialization of additional KV projections stabilizes fine-tuning by mitigating the introduction of harmful noise, thereby enabling a seamless integration of task-specific object-centric representations with general-purpose \ac{vla} pre-training. As a result, \method significantly reduces data requirements for task-specific adaptation, enhancing the efficiency of robotic manipulation deployment in the real world.

We demonstrate the efficacy and efficiency of \method across \textbf{8 diverse real-world tasks}, achieving robust performance \textbf{with only 10-20 demonstrations}. The evaluation tasks span diverse manipulation challenges: pick-and-place tasks with rigid, soft, and precision-critical objects, as well as complex manipulations including articulated object operation, object pouring, and deformable cloth folding. Empirically, \method attains an impressive \textbf{76.7\%} success rate across all tasks with very limited demonstrations, significantly surpassing baseline methods that achieve a mere 20.8\% success rate. Additionally, we demonstrate the extensibility of \method on \textbf{long-horizon tasks} and its robustness on \textbf{unseen objects and backgrounds}. Ablation studies confirm the necessity of three key components: (1) \ac{vla} pre-training for skill priors, (2) object-centric representation for efficient task grounding and learning, and (3) ControlNet-style conditioning for stable adaptation. In summary, \method bridges the gap between \textbf{large-scale \ac{vla} pre-training} and \textbf{efficient object-centric adaptation}, enabling robots to acquire complex skills from minimal demonstrations.

\section{Related Works}
\subsection{Few-shot Learning for Manipulation}
Reducing reliance on costly demonstrations while ensuring robust performance remains a key challenge in robotic manipulation. Early approaches synthesized training data through simulation-based augmentation~\citep{torne2024reconciling, mandlekar2023mimicgen} or combined imitation with reinforcement learning for robustness~\citep{mu2024robotwin}, yet these methods depend heavily on accurate object poses and CAD models, limiting real-world applicability. To overcome sim-to-real gaps, especially in contact-rich or deformable tasks, recent methods turn to \textit{human video priors}, using representations like R3M~\citep{nair2023r3m}, VIP~\citep{ma2022vip} or Ag2Manip~\cite{li2024ag2manip} to guide policy learning. However, these largely rely on 2D cues and struggle with precise spatial reasoning. Concurrently, few-shot learning aims to generalize from minimal demonstrations, with DenseMatcher~\citep{zhu2024densematcher} and SPOT~\citep{hsu2024spot} addressing transferable skill learning via 3D correspondences and object-centric planning. Yet, most approaches still require handcrafted grasping routines and accurate pose pipelines, and few leverage \textit{pre-trained \ac{vla}} as priors. Our work addresses these gaps by introducing ControlNet-style conditioning for object-centric policy modulation, enabling efficient few-shot adaptation in unstructured settings.

\subsection{Object-centric Representation Learning}
Object-centric representation decomposes complex scenes into manipulable entities, enabling more efficient reasoning for robot learning. Early methods represent objects by poses~\citep{tremblay2018deep, tyree20226, migimatsu2020object} or by bounding boxes~\citep{wang2019deep, devin2018deep}, but their dependence on known categories or instance labels limits generalization to novel objects and dynamic environments. Unsupervised discovery techniques segment visual inputs into object-like regions without requiring labels~\citep{locatello2020object, burgess2019monet}, yet they struggle in cluttered or occluded scenes and often produce inconsistent object identities~\citep{wang2021generalization, heravi2023visuomotor}. Furthermore, most existing approaches fail to leverage large-scale pre-trained models effectively, aligning instead to category-specific or pose-based features~\citep{didolkar2024zero, yoon2023investigation, gao2023sa6d, yi2022object}, which necessitates extensive task-specific tuning and undermines the benefits of transfer learning. To address these challenges, we introduce a unified framework that integrates object-centric decomposition with large‑scale, data‑driven representations, thereby improving adaptability and transferability in real-world manipulation tasks.

\subsection{ControlNet-style Fine-tuning}
ControlNet~\citep{zhang2023adding} enhances large-scale pre-trained Stable Diffusion~\citep{stability2022stable} by efficiently incorporating additional conditional inputs, such as sketch, normal map, depth map or human pose, through zero-initialized convolution layers. These layers start with zero weights and bias, ensuring no initial impact on outputs while progressively learning to integrate new conditions. This methodology has been extensively applied in various domains, such as controllable image generation~\citep{zhang2023adding,li2024controlnet++,zavadski2023controlnet}, video generation~\citep{guo2024animatediff, wang2024disco, bar2024lumiere}, and human motion generation~\citep{dai2024motionlcm, xie2024omnicontrol}. Despite its widespread adoption in these areas, the application of ControlNet in the context of robotic manipulation has not yet been investigated. Our study represents the first effort to adapt ControlNet-style fine-tuning to this field, unifying large-scale \ac{vla} pre-training with fine-tuning of object-centric representations to enable few-shot robotic manipulation learning. 

\section{Preliminary}
We formulate robot manipulation as a implicit discrete-time \ac{mdp} $\mathcal{M} = (\mbold{\mathcal{S}}, \mbold{\mathcal{A}}, \mbold{\mathcal{T}}, \mbold{\mathcal{R}}, \gamma)$, where $\mbold{\mathcal{S}}$ represents the state space, $\mbold{\mathcal{A}}$ the action space, $\mbold{\mathcal{T}}$ the stochastic transition probability, $\mbold{\mathcal{R}}$ the binary reward function with $\mbold{r} \in \left\{0, 1\right\}$, and $\gamma \in \left[0, 1\right)$ the discount factor. 
In our context, $\mbold{\mathcal{S}}$ denotes all robot and environment configurations, while $\mbold{\mathcal{A}}$ denotes the space of the robot's motor commands at each discrete time $t$. Our objective is to learn a closed-loop \ac{vla} $\mbold{\pi}: \mbold{\mathcal{O}} \rightarrow \mbold{\mathcal{A}}$~, where $\mbold{\mathcal{O}}$ is the observation space consisting of the robot's proprioception, RGB images, and language instruction, which serves as a partial projection of the state space $\mbold{\mathcal{S}}$ derived from the real-world sensors. We provide further preliminary of diffusion policy~\citep{chi2023diffusion} in the appendix. %

\begin{figure}[!t]
    \centering
    \includegraphics[width=\linewidth,]{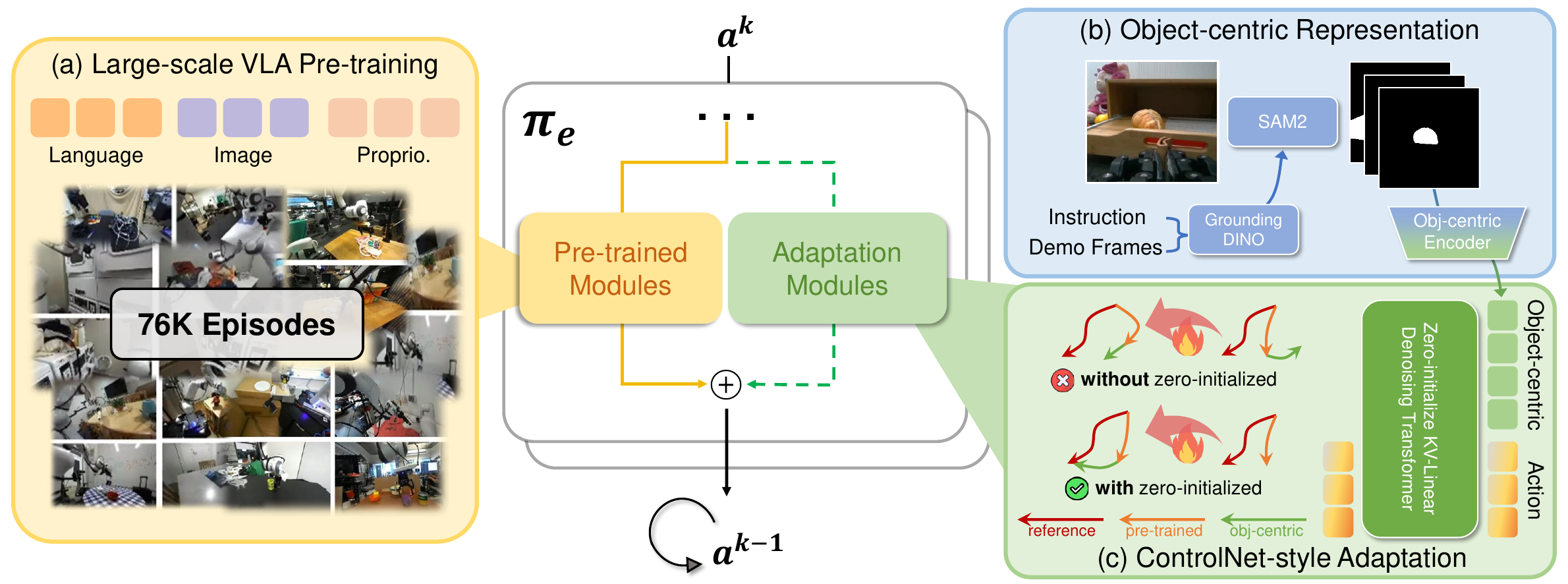}
    \caption{\textbf{Overview of \method.} \method leverages a ControlNet-style fine-tuning strategy to integrate object-centric representations with the pre-trained \ac{vla}. The zero-initialized weights and biases preserve the rich prior knowledge of the pre-trained policy while progressively grounding it in object-centric representation.}
    \label{fig:method}
    \vspace{-10pt}
\end{figure}

\section{Method}
Given a pre-trained general-purpose policy $\mbold{\pi_g}:\mbold{\mathcal{O}} \rightarrow \mbold{\mathcal{A}}$, our goal is to efficiently adapt it into a task-specific expert policy $\mbold{\pi_e}$ using limited expert demonstrations. To this end, we propose \method, which employs a ControlNet-style fine-tuning that integrates object-centric representations with a pre-trained \ac{vla} (see \cref{fig:method}). We first pre-train the general-purpose policy on a large-scale, multi-task manipulation dataset (\cref{sec:method:pretrain}), then extract object-centric features to focus learning on task-relevant elements (\cref{sec:method:object}), and finally fine-tune the policy by gradually incorporating these features (\cref{sec:method:control}). Implementation details are provided in the appendix. %

\subsection{VLA Model Pre-training}
\label{sec:method:pretrain}
We begin by pre-training a general-purpose policy $\mbold{\pi_g}$, using the public large-scale manipulation datasets $\mbold{\mathcal{D}_g} = \left\{ \left( \mbold{o_t}, \mbold{a_t} \right)_{t=1}^{T_i} \right\}_{i=1}^{N_g}$ across a diverse range of tasks and scenes, where $N_g$ represents the total number of episodes. Formally, we use $\mbold{\pi_g}: \mbold{\mathcal{O}} \rightarrow \mbold{\mathcal{A}}$ to model the conditional action distribution $p \left( \mbold{A_t} \mid \mbold{O_t} \right)$, where $\mbold{A_t}$ represents the future action sequence and $\mbold{O_t}$ denotes the history of the observations. The observation at time $t$ consists of a single-view RGB image $\mbold{I_t}$, a language instruction $\ell_t$, and the robot's proprioceptive state $\mbold{q_t}$, such that $\mbold{o_t} = \left[ \mbold{I_t}, \ell_t, \mbold{q_t} \right]$. The image $\mbold{I_t}$ and language instruction $\ell_t$ are tokenized via pre-trained encoders and projected into a shared embedding space through a linear projection layer, while the proprioceptive state $\mbold{q_t}$ is similarly embedded using \ac{mlps}.

The $\mbold{\pi_g}$ model adopts a diffusion transformer architecture to capture the multimodal conditional action distribution. During training, the action sequence is supervised using a conditional denoising loss. At inference, actions are sampled by iteratively denoising started from pure Gaussian noise $\mbold{A_t}^K \sim \mathcal{N}\left( 0, I \right)$ into the desired action $\mbold{A_t}^0 \sim q_\theta\left( \mbold{A_t}^0 \mid \mbold{O_t} \right)$, with the process accelerated using \ac{ddim}~\citep{song2021denoising} for real-time control.

\subsection{Object-centric Representations}
\label{sec:method:object}
This section outlines the process of building object-centric representations $\mbold{Z} \in \mbold{\mathcal{Z}}$ as additional action conditions, which enable task-specific expert policy $\mbold{\pi_e}$ to explicitly identify the key concepts of the task and efficiently learn from few-shot demonstrations. The process involves two key steps: (i) segment and track task-relevant objects, and (ii) learn to extract object-centric representations.

\textbf{Segment and Track task-relevant objects.}
To build object representations that consistently attend to task-relevant objects, it is essential to inform the model about their locations and local geometry in the RGB image observation $\mbold{I}$. Our goal is to automatically access fine-grained instance masks $\left\{\mbold{M^i}\right\}_{i=1}^{N_\mathrm{obj}}$ corresponding to task-relevant objects $\left\{\mbold{\mathrm{obj}^i}\right\}_{i=1}^{N_\mathrm{obj}}$, where ${N_\mathrm{obj}}$ denotes the number of objects. To achieve this, we extract ${N_\mathrm{img}}$ frames from demonstration and language instruction as prompts, and leverage GroundingDINO~\cite{liu2024grounding} and SAM2~\cite{ravi2024sam} to consistently segment and track task-relevant objects, both in the training data and during real-time inference.

\textbf{Learn to extract object-centric representations.}
We aim to learn a $\mbold{f_\varphi}$ to extract the object-centric representations $\mbold{z^i}$ from $i\mathrm{-th}$ object mask $\mbold{M^i}$ as additional conditions for the expert policy $\mbold{\pi_e}$. To encode \textit{where} and \textit{what} relevant objects are, we design \textit{positional feature} and \textit{geometrical feature} for each object. For positional feature $\mbold{z_\mathrm{pos}^i}$, we encode the mean coordinates of the object mask on images using sinusoidal positional encoding~\citep{vaswani2017attention}. For geometrical feature $\mbold{z_\mathrm{geo}^i}$, we obtain a spatial feature map with a \ac{cnn}~\citep{krizhevsky2012imagenet} that runs on the mask of each task-relevant object. Similar to the approach of Zhu~\etal~\citep{zhu2023viola}, we train the spatial network from scratch rather than using a pre-trained model, as we require actionable visual features that are specifically informative for continuous control tasks. 
Finally, we concatenate the positional and geometry feature to form the object-centric representation $\mbold{z^i} = \left[ \mbold{z_\mathrm{pos}^i}, \mbold{z_\mathrm{geo}^i} \right]$ and $\mbold{Z} = \left\{ \mbold{z^i} \right\}_{i=1}^{N_\mathrm{obj}} \in \mbold{\mathcal{Z}}$

\subsection{ControlNet-style Fine-tuning}
\label{sec:method:control}
Given a small set of task-specific dataset $\mbold{\mathcal{D}_e} = \left\{ \left( \mbold{o_t}, \mbold{z_t}, \mbold{a_t} \right)_{t=1}^{T_i} \right\}_{i=1}^{N_e}$, where $N_e$ represents the number of demonstrations, we aim to efficiently fine-tune an expert policy $\mbold{\pi_e}: \mbold{\mathcal{O}} \times \mbold{\mathcal{Z}} \rightarrow \mbold{\mathcal{A}}$ from $\mbold{\pi_g}: \mbold{\mathcal{O}} \rightarrow \mbold{\mathcal{A}}$ with the object-centric representations.

In our context, the pre-trained policy is a transformer-based architecture that utilizes cross-attention blocks to model actions conditioned on observations. Specifically, the cross-attention mechanism computes the relationship between actions $\mbold{A}$ and observations $\mbold{O}$ as:
\begin{equation}
\label{eq:attn}
    \mathrm{softmax}\left(\frac{\mbold{Q}\mbold{K}^T}{\sqrt{d}}\right)\mbold{V},
\end{equation}
where $\mbold{Q} = \mathbf{W_a} \mbold{A}  + \mathbf{B_a}$ represents the query projection, and $\mbold{K}, \mbold{V} = \mathbf{W_o} \mbold{O} + \mathbf{B_o}$ represent the key and value projections, respectively. To incorporate the object-centric representation $\mbold{Z} \in \mbold{\mathcal{Z}}$, we extend the cross-attention mechanism by introducing a dual-attention structure. 
\begin{equation}
\label{eq:attn_obj}
    \mathrm{softmax}\left(\frac{\mbold{Q}\mbold{K}^T}{\sqrt{d}}\right)\mbold{V} + \mathrm{softmax}\left(\frac{\mbold{Q}\mbold{K_z}^T}{\sqrt{d}}\right)\mbold{V_z},
\end{equation}
where $\mbold{K_z}, \mbold{V_z} = \mathbf{W_z} \mbold{Z} + \mathbf{B_z}$ are the key and value projections for the object-centric observations.

Inspired by Zhang~\etal~\citep{zhang2023adding}, we \textit{zero-initialize} the additional KV-projection layers to ensure the expert policy $\mbold{\pi_e}$ model behaves similarly to the pre-trained general-purpose policy $\mbold{\pi_g}$ during the early stage of fine-tuning, preserving the model's prior knowledge. 
Since the weight $\mathbf{W_z}$ and bias $\mathbf{B_z}$ are initialized to $\mathbf{0}$, the key and value projections for $\mbold{Z}$ are zero:
\begin{equation}
    \mathbf{K_z} = \mathbf{W_z} \mbold{Z} + \mathbf{B_z} = \mathbf{0}, \quad \mathbf{V_z} = \mathbf{0}.
\end{equation}
Thus, the dual-attention in \cref{eq:attn_obj} reduces to the original cross-attention in \cref{eq:attn}, preserving the pre-trained policy's behavior at the first fine-tuning step. We provide further explanation in the appendix. %

\section{Experiments}
To evaluate the efficiency of \method, we conduct 8 various real-world tasks using only 10-20 demonstrations. Empirically, our findings indicate that our method consistently and significantly improves success rates across all short-horizon tasks, achieving an overall success rate of \textbf{76.7\%}, which markedly surpasses the baseline of 20.8\%. For long-horizon tasks, we evaluate \method on two challenging scenarios, where it outperforms the state-of-the-art method with approximately 3x higher success rates. We further assess policy performance on data scaling. Our results show that our method rapidly converges to a high success rate with as few as 20 demonstrations, while baselines require more than 100 demonstrations to achieve comparable performance. Additionally, we also demonstrate the robustness of our method over unseen objects and backgrounds.

\subsection{Experimental Setup}
\begin{figure*}[!t]
    \centering
    \includegraphics[width=\linewidth]{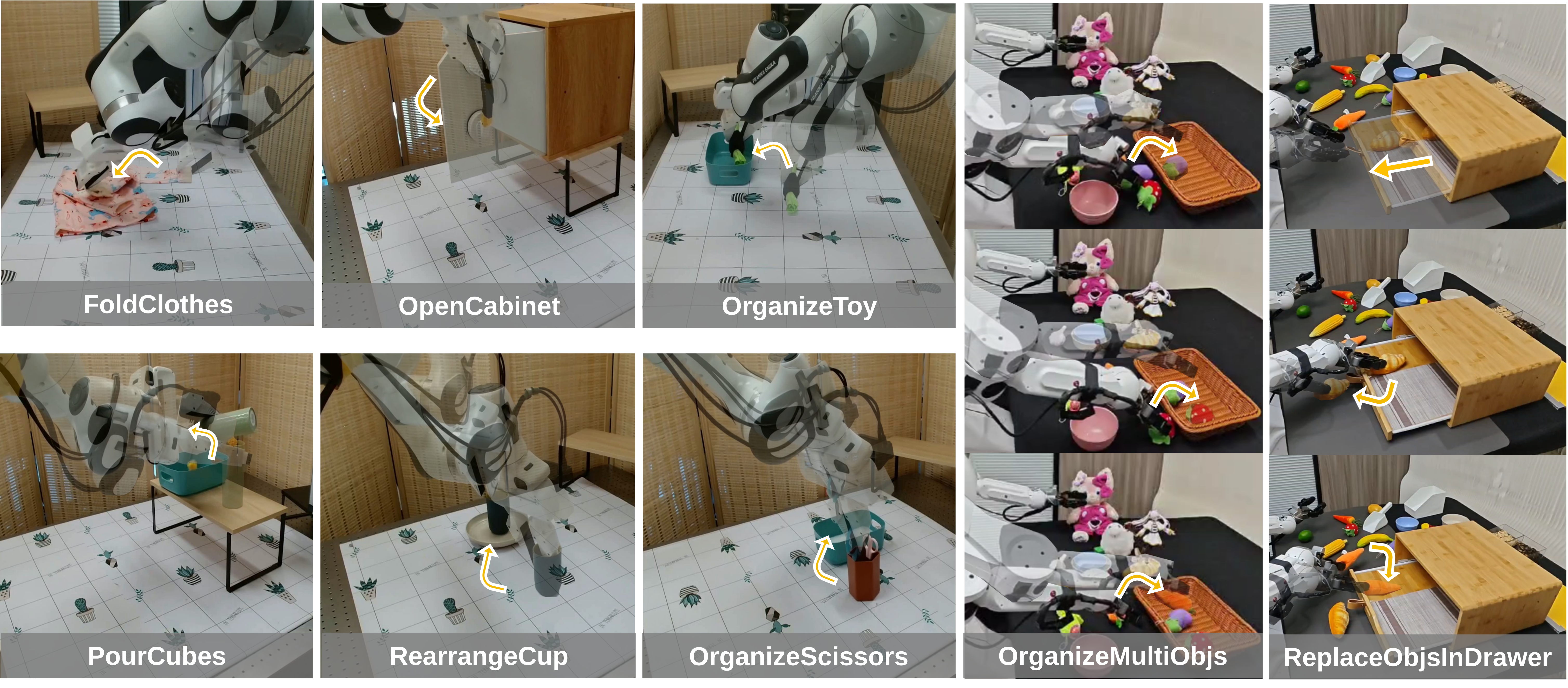}
    \vspace{-14pt}
    \caption{\textbf{Task Visualization.} The initial and target states are shown as transparent and solid layers, respectively. The yellow arrow highlights the desired transition.}
    \label{fig:task_visualization}
    \vspace{-12pt}
\end{figure*}

\textbf{Tasks.}
We develop a suite of various real-world tasks to evaluate the efficacy of our proposed method. These tasks are designed to cover a wide range of manipulation challenges, including pick-and-place various types of objects like rigid~\texttt{RearrangeCup}, soft~\texttt{OrganizeToy}, and precise~\texttt{OrganizeScissors}, as well as articulated~\texttt{OpenCabinet}, deformable~\texttt{FoldClothes}, and fluid-like~\texttt{PourCubes} manipulations. To further evaluate the performance in long-horizon scenarios, we introduce 2 additional tasks: \texttt{OrganizeMultiObjs} and \texttt{ReplaceObjInDrawer}, which entail sequential decision-making and handling of temporally extended goals. A detailed illustration of each task is given in~\cref{tab:task_illustration} while visualization is provided in~\cref{fig:task_visualization}.

\begin{table*}[!t]
\centering
\caption{\textbf{Illustrations of Evaluation Tasks.} We develop a suite real-world tasks for evaluation, including rigid, soft, articulated, deformable, and fluid-like objects. Two tasks are long-horizon, involving temporally extended behaviors with multiple sub-goals and sustained policy control.}
\vspace{-6pt}
\label{tab:task_illustration}
\resizebox{\linewidth}{!}{%
    \begin{tabular}{cc@{}c@{}c}
    \toprule
    \textbf{TASK NAME} & \textbf{TASK TYPE} & \textbf{\#DEMOS} & \textbf{LANGUAGE DESCRIPTION} \\ 
    \midrule
    \texttt{RearrangeCup} & Rigid Pick\&Place & 14 & Rearrange the cup and set it on the light plate. \\ 
    \texttt{OrganizeToy} & Soft Pick\&Place & 20 & Pick up the green toy and place it in the blue bowl. \\ 
    \texttt{OrganizeScissors} & Precise Pick\&Place & 15 & Pick up the scissors from the pen holder into the blue basket. \\ 
    \texttt{OpenCabinet} & Articulated Manipulation & 11 & Open the cabinet with the black handle. \\ 
    \texttt{FoldClothes} & Deformable Manipulation & 16 & Fold up the sleeves of the pink clothing on the table. \\ 
    \texttt{PourCubes} & Pouring Behavior & 19 & Pour the blocks from the green cup into the blue box. \\ 
    \texttt{OrganizeMultiObjs} & Long Horizon & 25 & Organize eggplant, strawberry, and carrot into the woven basket. \\
    \texttt{ReplaceObjInDrawer} & Long Horizon & 25 & Open the drawer, take out the bread, and put the carrot in. \\ 
    \bottomrule
    \end{tabular}
}%
\vspace{-12pt}
\end{table*}

\textbf{Baselines and Ablations.}
We compare our method against Octo~\citep{team2024octo}, $\pi_0$~\cite{black2410pi0}, VIOLA~\citep{zhu2023viola}, ACT~\citep{zhao2023learning}, and Diffusion Policy~\citep{chi2023diffusion}. Octo and $\pi_0$ are pre-trained foundation \ac{vla}, employing a diffusion-based model to decode the action tokens. VIOLA is a widely recognized 2D object-centric transformer-based policy learning framework that utilizes \ac{rpn}~\citep{zhou2022detecting} to extract object-centric representations. ACT and Diffusion Policy are among the most extensively studied and widely applied imitation visuomotor policies. ACT models the actions using a \ac{vae}, while Diffusion Policy leverages \ac{ddpm} to capture more multimodal action distributions. 

In our ablation study, we systematically remove individual components from our method to investigate their independent contributions. We ablate the pre-training phase by training an object-centric Diffusion Policy from scratch, denoted as ``w/o pretrain''. We eliminate the object-centric representations by directly fine-tuning the pre-trained model, denoted as ``w/o object-centric''. To assess the significance of ControlNet-style fine-tuning in integrating object-centric representations into pre-trained policies, we omit the zero-initialization of the projection layers for additional object-centric conditions, denoted as ``w/o zero-init''.

We provide \textbf{Data Collection} and \textbf{Evaluation Setup and Protocol} details in the appendix.

\begin{figure}[!t]
    \centering
    \includegraphics[width=\linewidth]{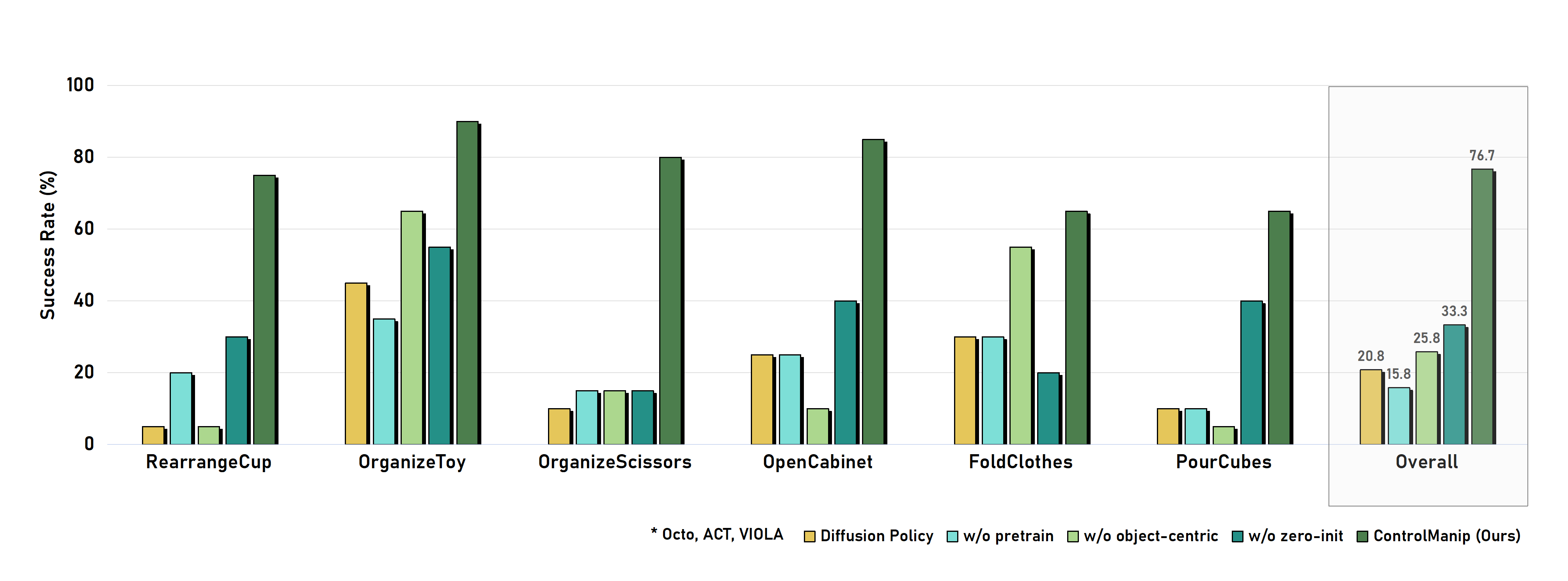}
    \vspace{-20pt}
    \caption{\textbf{Main Comparison and Ablation Study.} All policies are trained or fine-tuned from a shared, limited demonstration dataset for each task. *Octo, ACT, and VIOLA are omitted due to very low success rates, with overall success rates of 1.6\%, 5.0\%, and 0.0\%, respectively.}
    \label{fig:main_comparison}
    \vspace{-10pt}
\end{figure}

\begin{table}[t]
    \begin{minipage}[t]{0.76\linewidth}
        \centering
        \small
        \setlength{\tabcolsep}{2pt}
        \caption{\textbf{Performance on Long-horizon Tasks.}}
        \label{tab:long_horizon}
        \begin{tabular}{lcccclccccl}
            \toprule
            \multirow{2}{*}{} &
            \multicolumn{4}{c}{\texttt{OrganizeMultiObjects}} && 
            \multicolumn{4}{c}{\texttt{ReplaceObjectsInDrawer}} \\
            \cmidrule{2-5} \cmidrule{7-10}
            & Stage-1 & Stage-2 & Stage-3 & Overall && Stage-1 & Stage-2 & Stage-3 & Overall \\
            \midrule
            DP~\cite{chi2023diffusion} & 14$/$30 & 7$/$14  & 3$/$7   & 10.0\% && 11$/$30 & 5$/$11 & 2$/$5  & 6.7\%  \\
            $\pi_0$~\cite{black2410pi0} & 18$/$30 & 10$/$18 & 7$/$10  & 23.3\% && 19$/$30 & 9$/$19 & 5$/$9  & 16.7\% \\
            \method & 26$/$30 & 23$/$26 & 17$/$23 & \textbf{56.7\%} && 25$/$30 & 22$/$25 & 19$/$20 & \textbf{63.3\%} \\
            \bottomrule
        \end{tabular}
    \end{minipage}
    \hfill
    \begin{minipage}[t]{0.22\linewidth}
        \vspace{3pt}
        \centering
        \small
        \caption{Unseen Test.}
        \label{tab:unseen_domain}
        \begin{tabular}{lc}
            \toprule
             & \textbf{SR} \\
            \midrule
            obj-1 & 76.7\% \\
            obj-2 & 63.3\% \\
            obj-3 & 70.0\% \\
            bg   & 60.0\% \\
            \bottomrule
        \end{tabular}
    \end{minipage}
    \vspace{-8pt}
\end{table}

\subsection{Comparative and Ablation Results}
For full comparison and ablation studies, we evaluate each model over 20 trials across 6 short-horizon tasks. All policies are trained or fine-tuned from a shared, limited demonstration dataset for each task. As illustrated in \cref{fig:main_comparison}, the task success rates are presented within and across all evaluation tasks, providing a comprehensive overview of our findings. 

Our method, \method, achieves an impressive \textbf{overall task success rate of 76.7\%}, significantly outperforming the baselines. The strongest baseline, Diffusion Policy, attains only a 20.8\% success rate and struggles to precisely manipulate target objects or learn complex behaviors such as pouring. Octo and ACT only achieve 2$/$20 and 6$/$20 success on the \texttt{OpenCabinet} task, with no successes in other tasks, resulting in the overall success rate of just 1.6\% and 5.0\%.  Despite Octo's advantage from large-scale pre-training, its regression-based backbone fails to model action distributions with multiple distinct modes. While ACT leverages a \ac{vae} to represent action diversity, it is hindered by posterior collapse, especially in the low-data regime. VIOLA fails primarily due to the lack of action pre-training and its object-centric representation extraction, which relies on a large number of task demonstrations. In contrast, \method demonstrates robust and efficient learning across various real-world tasks even when only limited demonstrations are available.

Ablation studies reveal the critical role of key components of \method. Removing the pre-training phase and directly incorporating object-centric representations (``w/o pretrain") results in severe jitter problems when policies deploy to a real robot, causes severe execution jitter and significant drops in performance across tasks (details are provided in the appendix). We hypothesize that the object-centric features may provide deceptive low-loss pathways during the early training stage, incentivizing the policy to bypass learning from the more stable visual features. Eliminating object-centric representations during fine-tuning (``w/o object-centric") provides only a marginal improvement over training Diffusion Policy from scratch, highlighting the importance of object-centric representations. Finally, removing zero-initialization for additional object-centric conditions (``w/o zero-init") causes a drastic drop in success rate, emphasizing the role of proper initialization in stabilizing training and improving task performance. Notably, our complete methodology degenerates to the Diffusion Policy baseline when stripping all three components (pre-training, object-centric representations, and zero-initialization).

\subsection{Performance on Long-horizon Tasks}
We evaluate the ability of \method to perform long-horizon tasks, each trained with only 25 demonstrations. These tasks require the robot to execute multiple sub-goals in sequence, making them particularly challenging under limited supervision. Specifically, \texttt{OrganizeMultiObjects} involves placing three objects—eggplant, strawberry, and carrot—into a woven basket, corresponding to three sequential stages. \texttt{ReplaceObjectsInDrawer} requires opening a drawer, taking out a bread item, and placing a carrot inside, similarly decomposable into three dependent stages.

As shown in \cref{tab:long_horizon}, \method significantly outperforms baselines under this low-data regime, achieving an average 60.0\% success rate on two long-horizon tasks. The stage-wise breakdown further demonstrates \method's consistent performance throughout long action sequences. These results highlight the robustness of \method in long-horizon settings, even with minimal demonstrations, and its ability to reduce compounding errors during sequential execution.

\begin{wrapfigure}{!r}{0.40\linewidth} 
    \centering
    \vspace{-40pt}
    \includegraphics[width=1\linewidth]{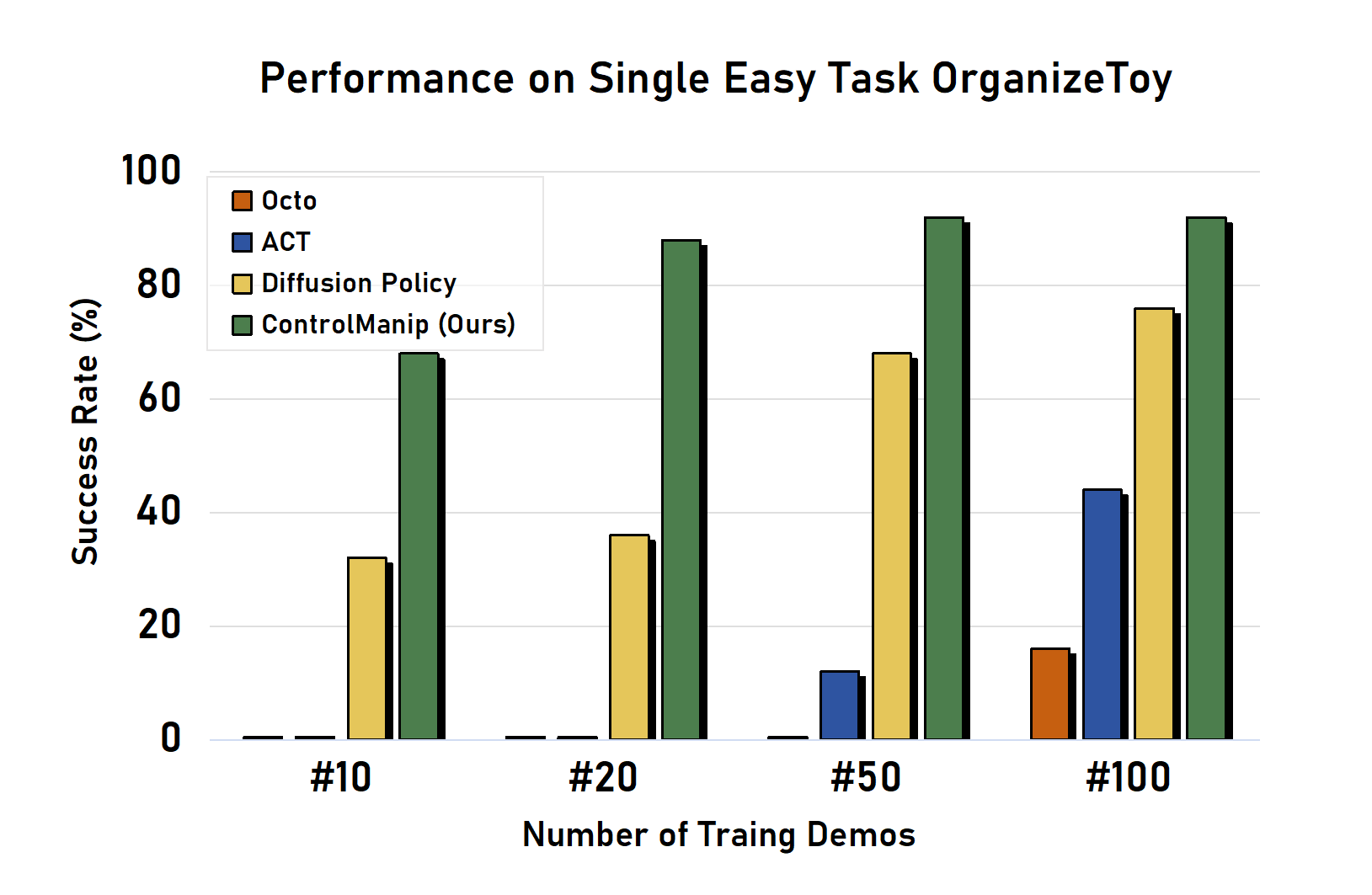}
    \vspace{-20pt}
    \caption{Effect of Data Scaling on Performance in the \texttt{OrganizeToy} Task.}
    \label{fig:data_scaling}
    \vspace{10pt}
\end{wrapfigure}

\subsection{Performance on Data Scaling}
We evaluate \method's data efficiency through controlled scaling experiments on the \texttt{OrganizeToy} task, benchmarking against established baseline methods. Each approach is tested across demonstration set sizes of 10, 20, 50, and 100 episodes with 25 trials per condition, with results shown in~\cref{fig:data_scaling}. While all methods improve as the amount of training data increases, \method achieves a high success rate of 80\% with only 20 demonstrations --- a level unattained by baselines even at 100 demonstrations. This highlights the efficiency of \method in learning from limited data.

\begin{wrapfigure}{!r}{0.40\linewidth} 
    \centering
    \vspace{-74pt}
    \includegraphics[width=0.9\linewidth]{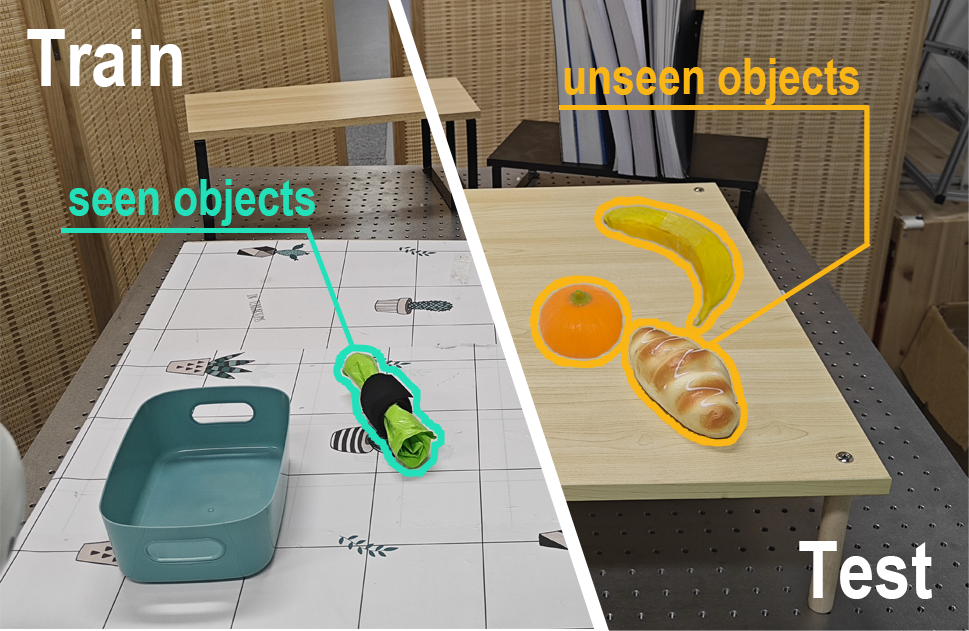}
    \caption{Generalization over object and background appearance changes.}
    \label{fig:unseen_domain}
    \vspace{-10pt}
\end{wrapfigure}

\subsection{Generalization on Object and Background}
We evaluate the generalization and robustness of \method on the \texttt{OrganizeToy} task by testing it with unseen objects and backgrounds. Trained on 20 demonstrations using a green toy and uniform background, \method achieves a 70.0\% average success rate across three novel objects (bread, banana, orange) and a 60.0\% success rate on a previously unseen background—requiring only an object prompt for mask extraction (\cref{tab:unseen_domain}). Although these results fall short of the 90\% in-domain success rate, they highlight \method's capacity to adapt to dynamic and diverse environments without additional training.

\section{Conclusion} 
This work introduces \method, a framework that bridges \ac{vla} pre-training with object-centric representations to enable efficient few-shot adaptation for robotic manipulation. By integrating a ControlNet-style fine-tuning, \method injects task-specific object-centric conditions into a pre-trained \ac{vla} through zero-initialized key-value projection layers. This design preserves the rich prior knowledge of the base policy while progressively grounding it in structured object properties, achieving stable fine-tuning with minimal demonstrations. 
Across 8 diverse real-world tasks—ranging from rigid object pick-and-place to deformable cloth folding and long-horizon tasks --- \method achieves a 76.7\% success rate with only 10–20 demonstrations, outperforming the baselines. By reducing demonstration requirements to practical levels, \method lowers barriers to deploying robots in diverse scenarios.

\section{Limitations}
Our evaluation tasks are currently constrained to single-arm manipulation, with experiments conducted primarily in controlled indoor environments. While these settings provide a reliable testbed, they fall short of capturing the full range of challenges faced in real-world deployment. To improve the generalization of \method, it is essential to explore bi-manual and in-the-wild manipulation tasks. Nevertheless, \method is a general framework, and future work could explore larger pre-trained bi-manual foundation \ac{vla}, along with more efficient and task-relevant modalities and representations to tackle these challenges.

\acknowledgments{We thank Yuyang Li (BIGAI, PKU) for his technical support and contributed discussion, Yuwei Guo (CUHK) for his discussion, and Ziyuan Jiao (BIGAI) for his assistance in setting up the real-world environment.}

\bibliography{references}  %

\begin{thebibliography}{63}
\providecommand{\natexlab}[1]{#1}
\providecommand{\url}[1]{\texttt{#1}}
\expandafter\ifx\csname urlstyle\endcsname\relax
  \providecommand{\doi}[1]{doi: #1}\else
  \providecommand{\doi}{doi: \begingroup \urlstyle{rm}\Url}\fi

\bibitem[Fu et~al.(2024)Fu, Zhao, and Finn]{fu2024mobile}
Z.~Fu, T.~Z. Zhao, and C.~Finn.
\newblock Mobile aloha: Learning bimanual mobile manipulation with low-cost whole-body teleoperation.
\newblock \emph{arXiv preprint arXiv:2401.02117}, 2024.

\bibitem[Li et~al.(2024)Li, Liu, Li, Han, Geng, Wang, Zhu, Zhu, and Huang]{li2024ag2manip}
P.~Li, T.~Liu, Y.~Li, M.~Han, H.~Geng, S.~Wang, Y.~Zhu, S.-C. Zhu, and S.~Huang.
\newblock Ag2manip: Learning novel manipulation skills with agent-agnostic visual and action representations.
\newblock In \emph{2024 IEEE/RSJ International Conference on Intelligent Robots and Systems (IROS)}, pages 573--580. IEEE, 2024.

\bibitem[Liu et~al.(2024)Liu, Wu, Li, Tan, Chen, Wang, Xu, Su, and Zhu]{liu2024rdt}
S.~Liu, L.~Wu, B.~Li, H.~Tan, H.~Chen, Z.~Wang, K.~Xu, H.~Su, and J.~Zhu.
\newblock Rdt-1b: a diffusion foundation model for bimanual manipulation.
\newblock \emph{arXiv preprint arXiv:2410.07864}, 2024.

\bibitem[Zhu et~al.(2023)Zhu, Joshi, Stone, and Zhu]{zhu2023viola}
Y.~Zhu, A.~Joshi, P.~Stone, and Y.~Zhu.
\newblock Viola: Imitation learning for vision-based manipulation with object proposal priors.
\newblock In \emph{Conference on Robot Learning}, pages 1199--1210. PMLR, 2023.

\bibitem[Chen et~al.(2024)Chen, Mu, Liang, Chen, Peng, Chen, Xu, Hu, Zhang, Li, et~al.]{chen2024g3flow}
T.~Chen, Y.~Mu, Z.~Liang, Z.~Chen, S.~Peng, Q.~Chen, M.~Xu, R.~Hu, H.~Zhang, X.~Li, et~al.
\newblock G3flow: Generative 3d semantic flow for pose-aware and generalizable object manipulation.
\newblock \emph{arXiv preprint arXiv:2411.18369}, 2024.

\bibitem[Wang et~al.(2024)Wang, Yin, Huang, Kelestemur, Wang, and Li]{wang2024gendp}
Y.~Wang, G.~Yin, B.~Huang, T.~Kelestemur, J.~Wang, and Y.~Li.
\newblock Gendp: 3d semantic fields for category-level generalizable diffusion policy.
\newblock In \emph{8th Annual Conference on Robot Learning}, volume~2, 2024.

\bibitem[Hsu et~al.(2024)Hsu, Wen, Xu, Narang, Wang, Zhu, Biswas, and Birchfield]{hsu2024spot}
C.-C. Hsu, B.~Wen, J.~Xu, Y.~Narang, X.~Wang, Y.~Zhu, J.~Biswas, and S.~Birchfield.
\newblock Spot: Se (3) pose trajectory diffusion for object-centric manipulation.
\newblock \emph{arXiv preprint arXiv:2411.00965}, 2024.

\bibitem[Chi et~al.(2024)Chi, Xu, Pan, Cousineau, Burchfiel, Feng, Tedrake, and Song]{chi2024universal}
C.~Chi, Z.~Xu, C.~Pan, E.~Cousineau, B.~Burchfiel, S.~Feng, R.~Tedrake, and S.~Song.
\newblock Universal manipulation interface: In-the-wild robot teaching without in-the-wild robots.
\newblock \emph{arXiv preprint arXiv:2402.10329}, 2024.

\bibitem[Ma et~al.(2023)Ma, Sodhani, Jayaraman, Bastani, Kumar, and Zhang]{ma2022vip}
Y.~J. Ma, S.~Sodhani, D.~Jayaraman, O.~Bastani, V.~Kumar, and A.~Zhang.
\newblock Vip: Towards universal visual reward and representation via value-implicit pre-training.
\newblock In \emph{The Eleventh International Conference on Learning Representations}, 2023.

\bibitem[Jiang et~al.(2024)Jiang, Xie, Lin, Xu, Wan, Mandlekar, Fan, and Zhu]{jiang2024dexmimicgen}
Z.~Jiang, Y.~Xie, K.~Lin, Z.~Xu, W.~Wan, A.~Mandlekar, L.~Fan, and Y.~Zhu.
\newblock Dexmimicgen: Automated data generation for bimanual dexterous manipulation via imitation learning.
\newblock \emph{arXiv preprint arXiv:2410.24185}, 2024.

\bibitem[Brohan et~al.(2023)Brohan, Brown, Carbajal, Chebotar, Chen, Choromanski, Ding, Driess, Dubey, Finn, et~al.]{brohan2023rt}
A.~Brohan, N.~Brown, J.~Carbajal, Y.~Chebotar, X.~Chen, K.~Choromanski, T.~Ding, D.~Driess, A.~Dubey, C.~Finn, et~al.
\newblock Rt-2: Vision-language-action models transfer web knowledge to robotic control.
\newblock \emph{arXiv preprint arXiv:2307.15818}, 2023.

\bibitem[Yang et~al.(2024)Yang, Du, Ghasemipour, Tompson, Kaelbling, Schuurmans, and Abbeel]{yang2024learning}
S.~Yang, Y.~Du, S.~K.~S. Ghasemipour, J.~Tompson, L.~P. Kaelbling, D.~Schuurmans, and P.~Abbeel.
\newblock Learning interactive real-world simulators.
\newblock In \emph{The Twelfth International Conference on Learning Representations}, 2024.

\bibitem[Li et~al.(2025)Li, Li, Liu, Li, and Huang]{li2025maniptrans}
K.~Li, P.~Li, T.~Liu, Y.~Li, and S.~Huang.
\newblock Maniptrans: Efficient dexterous bimanual manipulation transfer via residual learning.
\newblock In \emph{Proceedings of the IEEE conference on computer vision and pattern recognition}, 2025.

\bibitem[Huang et~al.(2024)Huang, Yong, Ma, Linghu, Li, Wang, Li, Zhu, Jia, and Huang]{huang2024embodied}
J.~Huang, S.~Yong, X.~Ma, X.~Linghu, P.~Li, Y.~Wang, Q.~Li, S.-C. Zhu, B.~Jia, and S.~Huang.
\newblock An embodied generalist agent in 3d world.
\newblock In \emph{Proceedings of the 41st International Conference on Machine Learning}, pages 20413--20451, 2024.

\bibitem[Li et~al.(2023)Li, Liu, Li, Geng, Zhu, Yang, and Huang]{li2023gendexgrasp}
P.~Li, T.~Liu, Y.~Li, Y.~Geng, Y.~Zhu, Y.~Yang, and S.~Huang.
\newblock Gendexgrasp: Generalizable dexterous grasping.
\newblock In \emph{2023 IEEE International Conference on Robotics and Automation (ICRA)}, pages 8068--8074. IEEE, 2023.

\bibitem[Li et~al.(2024)Li, Liu, Geng, Li, Yang, Zhu, Liu, and Huang]{li2024grasp}
Y.~Li, B.~Liu, Y.~Geng, P.~Li, Y.~Yang, Y.~Zhu, T.~Liu, and S.~Huang.
\newblock Grasp multiple objects with one hand.
\newblock \emph{IEEE Robotics and Automation Letters}, 2024.

\bibitem[Torne et~al.(2024)Torne, Simeonov, Li, Chan, Chen, Gupta, and Agrawal]{torne2024reconciling}
M.~Torne, A.~Simeonov, Z.~Li, A.~Chan, T.~Chen, A.~Gupta, and P.~Agrawal.
\newblock Reconciling reality through simulation: A real-to-sim-to-real approach for robust manipulation.
\newblock \emph{arXiv preprint arXiv:2403.03949}, 2024.

\bibitem[Mandlekar et~al.(2023)Mandlekar, Nasiriany, Wen, Akinola, Narang, Fan, Zhu, and Fox]{mandlekar2023mimicgen}
A.~Mandlekar, S.~Nasiriany, B.~Wen, I.~Akinola, Y.~Narang, L.~Fan, Y.~Zhu, and D.~Fox.
\newblock Mimicgen: A data generation system for scalable robot learning using human demonstrations.
\newblock \emph{arXiv preprint arXiv:2310.17596}, 2023.

\bibitem[Mu et~al.(2024)Mu, Chen, Peng, Chen, Gao, Zou, Lin, Xie, and Luo]{mu2024robotwin}
Y.~Mu, T.~Chen, S.~Peng, Z.~Chen, Z.~Gao, Y.~Zou, L.~Lin, Z.~Xie, and P.~Luo.
\newblock Robotwin: Dual-arm robot benchmark with generative digital twins (early version).
\newblock \emph{arXiv preprint arXiv:2409.02920}, 2024.

\bibitem[Chi et~al.(2023)Chi, Xu, Feng, Cousineau, Du, Burchfiel, Tedrake, and Song]{chi2023diffusion}
C.~Chi, Z.~Xu, S.~Feng, E.~Cousineau, Y.~Du, B.~Burchfiel, R.~Tedrake, and S.~Song.
\newblock Diffusion policy: Visuomotor policy learning via action diffusion.
\newblock \emph{The International Journal of Robotics Research}, page 02783649241273668, 2023.

\bibitem[Nair et~al.(2023)Nair, Rajeswaran, Kumar, Finn, and Gupta]{nair2023r3m}
S.~Nair, A.~Rajeswaran, V.~Kumar, C.~Finn, and A.~Gupta.
\newblock R3m: A universal visual representation for robot manipulation.
\newblock In \emph{Conference on Robot Learning}, pages 892--909. PMLR, 2023.

\bibitem[Team et~al.(2024)Team, Ghosh, Walke, Pertsch, Black, Mees, Dasari, Hejna, Kreiman, Xu, et~al.]{team2024octo}
O.~M. Team, D.~Ghosh, H.~Walke, K.~Pertsch, K.~Black, O.~Mees, S.~Dasari, J.~Hejna, T.~Kreiman, C.~Xu, et~al.
\newblock Octo: An open-source generalist robot policy.
\newblock \emph{arXiv preprint arXiv:2405.12213}, 2024.

\bibitem[Kim et~al.(2024)Kim, Pertsch, Karamcheti, Xiao, Balakrishna, Nair, Rafailov, Foster, Lam, Sanketi, et~al.]{kim2024openvla}
M.~J. Kim, K.~Pertsch, S.~Karamcheti, T.~Xiao, A.~Balakrishna, S.~Nair, R.~Rafailov, E.~Foster, G.~Lam, P.~Sanketi, et~al.
\newblock Openvla: An open-source vision-language-action model.
\newblock \emph{arXiv preprint arXiv:2406.09246}, 2024.

\bibitem[Brohan et~al.(2022)Brohan, Brown, Carbajal, Chebotar, Dabis, Finn, Gopalakrishnan, Hausman, Herzog, Hsu, et~al.]{brohan2022rt}
A.~Brohan, N.~Brown, J.~Carbajal, Y.~Chebotar, J.~Dabis, C.~Finn, K.~Gopalakrishnan, K.~Hausman, A.~Herzog, J.~Hsu, et~al.
\newblock Rt-1: Robotics transformer for real-world control at scale.
\newblock \emph{arXiv preprint arXiv:2212.06817}, 2022.

\bibitem[O'Neill et~al.(2023)O'Neill, Rehman, Gupta, Maddukuri, Gupta, Padalkar, Lee, Pooley, Gupta, Mandlekar, et~al.]{o2023open}
A.~O'Neill, A.~Rehman, A.~Gupta, A.~Maddukuri, A.~Gupta, A.~Padalkar, A.~Lee, A.~Pooley, A.~Gupta, A.~Mandlekar, et~al.
\newblock Open x-embodiment: Robotic learning datasets and rt-x models.
\newblock \emph{arXiv preprint arXiv:2310.08864}, 2023.

\bibitem[Black et~al.(2024)Black, Brown, Driess, Esmail, Equi, Finn, Fusai, Groom, Hausman, Ichter, et~al.]{black2410pi0}
K.~Black, N.~Brown, D.~Driess, A.~Esmail, M.~Equi, C.~Finn, N.~Fusai, L.~Groom, K.~Hausman, B.~Ichter, et~al.
\newblock $\pi$0: A vision-language-action flow model for general robot control, 2024.
\newblock \emph{URL https://arxiv. org/abs/2410.24164}, 2024.

\bibitem[Zhu et~al.(2023)Zhu, Jiang, Stone, and Zhu]{zhu2023learning}
Y.~Zhu, Z.~Jiang, P.~Stone, and Y.~Zhu.
\newblock Learning generalizable manipulation policies with object-centric 3d representations.
\newblock In \emph{Conference on Robot Learning}, pages 3418--3433. PMLR, 2023.

\bibitem[Zhang et~al.(2023)Zhang, Rao, and Agrawala]{zhang2023adding}
L.~Zhang, A.~Rao, and M.~Agrawala.
\newblock Adding conditional control to text-to-image diffusion models.
\newblock In \emph{Proceedings of the IEEE/CVF International Conference on Computer Vision}, pages 3836--3847, 2023.

\bibitem[Zhu et~al.(2024)Zhu, Ju, Zhang, Wang, Yuan, Hu, and Xu]{zhu2024densematcher}
J.~Zhu, Y.~Ju, J.~Zhang, M.~Wang, Z.~Yuan, K.~Hu, and H.~Xu.
\newblock Densematcher: Learning 3d semantic correspondence for category-level manipulation from a single demo.
\newblock \emph{arXiv preprint arXiv:2412.05268}, 2024.

\bibitem[Tremblay et~al.(2018)Tremblay, To, Sundaralingam, Xiang, Fox, and Birchfield]{tremblay2018deep}
J.~Tremblay, T.~To, B.~Sundaralingam, Y.~Xiang, D.~Fox, and S.~Birchfield.
\newblock Deep object pose estimation for semantic robotic grasping of household objects.
\newblock \emph{arXiv preprint arXiv:1809.10790}, 2018.

\bibitem[Tyree et~al.(2022)Tyree, Tremblay, To, Cheng, Mosier, Smith, and Birchfield]{tyree20226}
S.~Tyree, J.~Tremblay, T.~To, J.~Cheng, T.~Mosier, J.~Smith, and S.~Birchfield.
\newblock 6-dof pose estimation of household objects for robotic manipulation: An accessible dataset and benchmark.
\newblock In \emph{2022 IEEE/RSJ International Conference on Intelligent Robots and Systems (IROS)}, pages 13081--13088. IEEE, 2022.

\bibitem[Migimatsu and Bohg(2020)]{migimatsu2020object}
T.~Migimatsu and J.~Bohg.
\newblock Object-centric task and motion planning in dynamic environments.
\newblock \emph{IEEE Robotics and Automation Letters}, 5\penalty0 (2):\penalty0 844--851, 2020.

\bibitem[Wang et~al.(2019)Wang, Devin, Cai, Yu, and Darrell]{wang2019deep}
D.~Wang, C.~Devin, Q.-Z. Cai, F.~Yu, and T.~Darrell.
\newblock Deep object-centric policies for autonomous driving.
\newblock In \emph{2019 International Conference on Robotics and Automation (ICRA)}, pages 8853--8859. IEEE, 2019.

\bibitem[Devin et~al.(2018)Devin, Abbeel, Darrell, and Levine]{devin2018deep}
C.~Devin, P.~Abbeel, T.~Darrell, and S.~Levine.
\newblock Deep object-centric representations for generalizable robot learning.
\newblock In \emph{2018 IEEE International Conference on Robotics and Automation (ICRA)}, pages 7111--7118. IEEE, 2018.

\bibitem[Locatello et~al.(2020)Locatello, Weissenborn, Unterthiner, Mahendran, Heigold, Uszkoreit, Dosovitskiy, and Kipf]{locatello2020object}
F.~Locatello, D.~Weissenborn, T.~Unterthiner, A.~Mahendran, G.~Heigold, J.~Uszkoreit, A.~Dosovitskiy, and T.~Kipf.
\newblock Object-centric learning with slot attention.
\newblock \emph{Advances in neural information processing systems}, 33:\penalty0 11525--11538, 2020.

\bibitem[Burgess et~al.(2019)Burgess, Matthey, Watters, Kabra, Higgins, Botvinick, and Lerchner]{burgess2019monet}
C.~P. Burgess, L.~Matthey, N.~Watters, R.~Kabra, I.~Higgins, M.~Botvinick, and A.~Lerchner.
\newblock Monet: Unsupervised scene decomposition and representation.
\newblock \emph{arXiv preprint arXiv:1901.11390}, 2019.

\bibitem[Wang et~al.(2021)Wang, Wang, Mandlekar, Fei-Fei, Savarese, and Xu]{wang2021generalization}
C.~Wang, R.~Wang, A.~Mandlekar, L.~Fei-Fei, S.~Savarese, and D.~Xu.
\newblock Generalization through hand-eye coordination: An action space for learning spatially-invariant visuomotor control.
\newblock In \emph{2021 IEEE/RSJ International Conference on Intelligent Robots and Systems (IROS)}, pages 8913--8920. IEEE, 2021.

\bibitem[Heravi et~al.(2023)Heravi, Wahid, Lynch, Florence, Armstrong, Tompson, Sermanet, Bohg, and Dwibedi]{heravi2023visuomotor}
N.~Heravi, A.~Wahid, C.~Lynch, P.~Florence, T.~Armstrong, J.~Tompson, P.~Sermanet, J.~Bohg, and D.~Dwibedi.
\newblock Visuomotor control in multi-object scenes using object-aware representations.
\newblock In \emph{2023 IEEE International Conference on Robotics and Automation (ICRA)}, pages 9515--9522. IEEE, 2023.

\bibitem[Didolkar et~al.(2024)Didolkar, Zadaianchuk, Goyal, Mozer, Bengio, Martius, and Seitzer]{didolkar2024zero}
A.~Didolkar, A.~Zadaianchuk, A.~Goyal, M.~Mozer, Y.~Bengio, G.~Martius, and M.~Seitzer.
\newblock Zero-shot object-centric representation learning.
\newblock \emph{arXiv preprint arXiv:2408.09162}, 2024.

\bibitem[Yoon et~al.(2023)Yoon, Wu, Bae, and Ahn]{yoon2023investigation}
J.~Yoon, Y.-F. Wu, H.~Bae, and S.~Ahn.
\newblock An investigation into pre-training object-centric representations for reinforcement learning.
\newblock \emph{arXiv preprint arXiv:2302.04419}, 2023.

\bibitem[Gao et~al.(2023)Gao, Ngo, Ziesche, and Neumann]{gao2023sa6d}
N.~Gao, V.~A. Ngo, H.~Ziesche, and G.~Neumann.
\newblock Sa6d: Self-adaptive few-shot 6d pose estimator for novel and occluded objects.
\newblock In \emph{7th Annual Conference on Robot Learning}, 2023.

\bibitem[Yi et~al.(2022)Yi, Zhang, Guo, Hu, Du, Guo, Chen, et~al.]{yi2022object}
Q.~Yi, R.~Zhang, J.~Guo, X.~Hu, Z.~Du, Q.~Guo, Y.~Chen, et~al.
\newblock Object-category aware reinforcement learning.
\newblock \emph{Advances in Neural Information Processing Systems}, 35:\penalty0 36453--36465, 2022.

\bibitem[Stability(2022)]{stability2022stable}
Stability.
\newblock Stable diffusion v1.5 model card.
\newblock \url{https://huggingface.co/runwayml/stable-diffusion-v1-5}, 2022.
\newblock Accessed: 2024-09-05.

\bibitem[Li et~al.(2024)Li, Yang, Kuang, Wu, Wang, Xiao, and Chen]{li2024controlnet++}
M.~Li, T.~Yang, H.~Kuang, J.~Wu, Z.~Wang, X.~Xiao, and C.~Chen.
\newblock Controlnet++: Improving conditional controls with efficient consistency feedback: Project page: liming-ai. github. io/controlnet\_plus\_plus.
\newblock In \emph{European Conference on Computer Vision}, pages 129--147. Springer, 2024.

\bibitem[Zavadski et~al.(2023)Zavadski, Feiden, and Rother]{zavadski2023controlnet}
D.~Zavadski, J.-F. Feiden, and C.~Rother.
\newblock Controlnet-xs: Designing an efficient and effective architecture for controlling text-to-image diffusion models.
\newblock \emph{arXiv preprint arXiv:2312.06573}, 2023.

\bibitem[Guo et~al.(2024)Guo, Yang, Rao, Liang, Wang, Qiao, Agrawala, Lin, and Dai]{guo2024animatediff}
Y.~Guo, C.~Yang, A.~Rao, Z.~Liang, Y.~Wang, Y.~Qiao, M.~Agrawala, D.~Lin, and B.~Dai.
\newblock Animatediff: Animate your personalized text-to-image diffusion models without specific tuning.
\newblock In \emph{The Twelfth International Conference on Learning Representations}, 2024.

\bibitem[Wang et~al.(2024)Wang, Li, Lin, Zhai, Lin, Yang, Zhang, Liu, and Wang]{wang2024disco}
T.~Wang, L.~Li, K.~Lin, Y.~Zhai, C.-C. Lin, Z.~Yang, H.~Zhang, Z.~Liu, and L.~Wang.
\newblock Disco: Disentangled control for realistic human dance generation.
\newblock In \emph{Proceedings of the IEEE/CVF Conference on Computer Vision and Pattern Recognition}, pages 9326--9336, 2024.

\bibitem[Bar-Tal et~al.(2024)Bar-Tal, Chefer, Tov, Herrmann, Paiss, Zada, Ephrat, Hur, Liu, Raj, et~al.]{bar2024lumiere}
O.~Bar-Tal, H.~Chefer, O.~Tov, C.~Herrmann, R.~Paiss, S.~Zada, A.~Ephrat, J.~Hur, G.~Liu, A.~Raj, et~al.
\newblock Lumiere: A space-time diffusion model for video generation.
\newblock In \emph{SIGGRAPH Asia 2024 Conference Papers}, pages 1--11, 2024.

\bibitem[Dai et~al.(2024)Dai, Chen, Wang, Liu, Dai, and Tang]{dai2024motionlcm}
W.~Dai, L.-H. Chen, J.~Wang, J.~Liu, B.~Dai, and Y.~Tang.
\newblock Motionlcm: Real-time controllable motion generation via latent consistency model.
\newblock In \emph{European Conference on Computer Vision}, pages 390--408. Springer, 2024.

\bibitem[Xie et~al.(2024)Xie, Jampani, Zhong, Sun, and Jiang]{xie2024omnicontrol}
Y.~Xie, V.~Jampani, L.~Zhong, D.~Sun, and H.~Jiang.
\newblock Omnicontrol: Control any joint at any time for human motion generation.
\newblock In \emph{The Twelfth International Conference on Learning Representations}, 2024.

\bibitem[Song et~al.(2021)Song, Meng, and Ermon]{song2021denoising}
J.~Song, C.~Meng, and S.~Ermon.
\newblock Denoising diffusion implicit models.
\newblock In \emph{International Conference on Learning Representations}, 2021.

\bibitem[Liu et~al.(2024)Liu, Zeng, Ren, Li, Zhang, Yang, Jiang, Li, Yang, Su, et~al.]{liu2024grounding}
S.~Liu, Z.~Zeng, T.~Ren, F.~Li, H.~Zhang, J.~Yang, Q.~Jiang, C.~Li, J.~Yang, H.~Su, et~al.
\newblock Grounding dino: Marrying dino with grounded pre-training for open-set object detection.
\newblock In \emph{European Conference on Computer Vision}, pages 38--55. Springer, 2024.

\bibitem[Ravi et~al.(2024)Ravi, Gabeur, Hu, Hu, Ryali, Ma, Khedr, R{\"a}dle, Rolland, Gustafson, et~al.]{ravi2024sam}
N.~Ravi, V.~Gabeur, Y.-T. Hu, R.~Hu, C.~Ryali, T.~Ma, H.~Khedr, R.~R{\"a}dle, C.~Rolland, L.~Gustafson, et~al.
\newblock Sam 2: Segment anything in images and videos.
\newblock \emph{arXiv preprint arXiv:2408.00714}, 2024.

\bibitem[Vaswani(2017)]{vaswani2017attention}
A.~Vaswani.
\newblock Attention is all you need.
\newblock \emph{Advances in Neural Information Processing Systems}, 2017.

\bibitem[Krizhevsky et~al.(2012)Krizhevsky, Sutskever, and Hinton]{krizhevsky2012imagenet}
A.~Krizhevsky, I.~Sutskever, and G.~E. Hinton.
\newblock Imagenet classification with deep convolutional neural networks.
\newblock \emph{Advances in neural information processing systems}, 25, 2012.

\bibitem[Zhao et~al.(2023)Zhao, Kumar, Levine, and Finn]{zhao2023learning}
T.~Z. Zhao, V.~Kumar, S.~Levine, and C.~Finn.
\newblock Learning fine-grained bimanual manipulation with low-cost hardware.
\newblock \emph{arXiv preprint arXiv:2304.13705}, 2023.

\bibitem[Zhou et~al.(2022)Zhou, Girdhar, Joulin, Kr{\"a}henb{\"u}hl, and Misra]{zhou2022detecting}
X.~Zhou, R.~Girdhar, A.~Joulin, P.~Kr{\"a}henb{\"u}hl, and I.~Misra.
\newblock Detecting twenty-thousand classes using image-level supervision.
\newblock In \emph{European Conference on Computer Vision}, pages 350--368. Springer, 2022.

\bibitem[Ho et~al.(2020)Ho, Jain, and Abbeel]{ho2020denoising}
J.~Ho, A.~Jain, and P.~Abbeel.
\newblock Denoising diffusion probabilistic models.
\newblock \emph{Advances in neural information processing systems}, 33:\penalty0 6840--6851, 2020.

\bibitem[Khazatsky et~al.(2024)Khazatsky, Pertsch, Nair, Balakrishna, Dasari, Karamcheti, Nasiriany, Srirama, Chen, Ellis, et~al.]{khazatsky2024droid}
A.~Khazatsky, K.~Pertsch, S.~Nair, A.~Balakrishna, S.~Dasari, S.~Karamcheti, S.~Nasiriany, M.~K. Srirama, L.~Y. Chen, K.~Ellis, et~al.
\newblock Droid: A large-scale in-the-wild robot manipulation dataset.
\newblock In \emph{Robotics: Science and Systems}, 2024.

\bibitem[Radford et~al.(2021)Radford, Kim, Hallacy, Ramesh, Goh, Agarwal, Sastry, Askell, Mishkin, Clark, et~al.]{radford2021learning}
A.~Radford, J.~W. Kim, C.~Hallacy, A.~Ramesh, G.~Goh, S.~Agarwal, G.~Sastry, A.~Askell, P.~Mishkin, J.~Clark, et~al.
\newblock Learning transferable visual models from natural language supervision.
\newblock In \emph{International conference on machine learning}, pages 8748--8763. PMLR, 2021.

\bibitem[Campos et~al.(2021)Campos, Elvira, Rodr{\'\i}guez, Montiel, and Tard{\'o}s]{campos2021orb}
C.~Campos, R.~Elvira, J.~J.~G. Rodr{\'\i}guez, J.~M. Montiel, and J.~D. Tard{\'o}s.
\newblock Orb-slam3: An accurate open-source library for visual, visual--inertial, and multimap slam.
\newblock \emph{IEEE transactions on robotics}, 37\penalty0 (6):\penalty0 1874--1890, 2021.

\bibitem[{Meta Platforms, Inc.}(n.d.)]{metaquest}
{Meta Platforms, Inc.}
\newblock {Meta Quest: Virtual Reality Headset}.
\newblock \url{https://www.meta.com/quest/}, n.d.
\newblock Accessed: 2025-05-08.

\bibitem[{Astribot, Inc.}(2024)]{astribot_s1}
{Astribot, Inc.}
\newblock {Astribot S1: AI Robotic Partner}.
\newblock \url{https://www.astribot.com/product-en}, 2024.
\newblock Accessed: 2025-05-06.

\end{thebibliography}

\clearpage
\appendix
\section{\texorpdfstring{More Experiments on $\pi_0$}{More Experiments on pi0}}
We further adapt and evaluate our ControlVLA method on a more general pre-trained \ac{vla}, $\pi_0$~\cite{black2410pi0}, referred to as ControlVLA@$\pi_0$. The $\pi_0$ model leverages a pre-trained vision-language backbone and introduces a separate \textit{action expert} that outputs continuous actions using flow matching. ControlVLA@$\pi_0$ extends this architecture by incorporating an additional \textit{object expert} and a set of zero-convolution layers to progressively inject object-centric representation guidance. These zero-convolution layers ensure that the object-centric cues are integrated as additional conditions without disrupting the learned action prior, enabling robust skill learning from limited demonstrations. We conduct the comparison studies over 20 trials across 4 sub-tasks, each fine-tuned with limited demonstrations, consistent with the setup used in the main paper.

As shown in~\cref{tab:result_pi0}, our adaptation method ControlVLA@$\pi_0$ consistently outperforms the fine-tuned $\pi_0$, demonstrating that ControlVLA can serve as a plug-in module to enhance performance across a broader range of pre-trained \ac{vla} models. The improvement is most pronounced on the \texttt{OrganizeScissors} task, highlighting ControlVLA's ability to provide more precise guidance for fine-grained manipulation in data-scarce scenarios.
\begin{table}[!h]
    \centering
    \caption{Task success rates of $\pi_0$ and ControlVLA@$\pi_0$ across various tasks.}
    \label{tab:result_pi0}
    \begin{tabular}{l *{4}{>{\centering\arraybackslash}m{40pt}} >{\centering\arraybackslash}m{35pt}}
        \toprule
         & Organize Toy
         & Organize Scissors
         & Open Cabinet
         & Fold Clothes
         & Overall \\
        \midrule
        $\pi_0$~\cite{black2410pi0} & 55.0\% & 15.0\% & 45.0\% & 40.0\% & 38.6\% \\
        ControlVLA@$\pi_0$ & 85.0\% & 80.0\% & 85.0\% & 75.0\% & \textbf{81.3\%} \\
        \bottomrule
    \end{tabular}
\end{table}

\section{Preliminary of Diffusion Policy}
\label{app:pre:dp}
Diffusion policy~\citep{chi2023diffusion} formulates the visuomotor policy $\mbold{\pi}$ as the \ac{ddpm}~\citep{ho2020denoising}, which can model complex multimodal action distributions and facilitate a stable training behavior. \ac{ddpm} performs $K$ iterations of a denoising process, starting from a Gaussian noise $\mbold{x}^K \sim \mathcal{N}\left(0, I\right)$ and evolving toward the desired output $\mbold{x}^0 \sim q_\theta\left(\mbold{x}^0\right)$. The denoising process is described by the following equation:
\begin{equation}
\label{eq:ddpm_process}
    \mbold{x}^{k-1} = \alpha \left(\mbold{x}^k - \beta \mbold{\epsilon_\theta}\left(\mbold{x}^k, k\right)\right) + \sigma \mathcal{N}\left(0, I\right),
\end{equation}
where $\alpha$, $\beta$, and $\sigma$ are functions of the timestep $k$, collectively known as the noise schedule, and the $\mbold{\epsilon_\theta}$ is the distribution shift prediction network with the trainable parameter $\theta$.

The training objective is to minimize the variational lower bound of KL-divergence between the given data distribution $p\left(\mbold{x}^0\right)$ and the $\theta$-parameterized distribution $q_\theta \left(\mbold{x}^0\right)$. As shown in \citep{ho2020denoising}, the loss function can be simplified as:
\begin{equation}
\label{eq:ddpm_loss}
    \mathcal{L} = \mathbb{E}_{t\sim\left[1, K\right], \mbold{x}^0, \epsilon^k} \left[ \| \epsilon^k - \mbold{\epsilon_\theta}\right(\mbold{x}^0 + \epsilon^k, k\left) \|^2 \right].
\end{equation}

Diffusion policy represents the robot actions $\mbold{a_{t:t+T_a}}$ as the model output $x$ and conditions the denoising process on the robot observations $\mbold{o_{t:t-T_o}}$, where $\mbold{a_t} \in \mbold{\mathcal{A}}$~, $\mbold{o_t} \in \mbold{\mathcal{O}}$~, $T_a$ and $T_o$ denote the horizon lengths of the action and observation sequences. For convenience, we use $\mbold{A_t}$ and $\mbold{O_t}$ to represent the action and observation sequences in the following discussion. The \ac{ddpm} is naturally extended to approximate the conditional distribution $p\left( \mbold{A_t} \mid \mbold{O_t} \right)$ for planning. To capture the conditional actions distribution, the denoising process is modified from \cref{eq:ddpm_process}:
\begin{equation}
\label{eq:ddpm_process_cond}
    \mbold{A_t}^{k-1} = \alpha \left(\mbold{A_t}^k - \beta \mbold{\epsilon_\theta}\left(\mbold{A_t}^k, k\right)\right) + \sigma \mathcal{N}\left(0, I\right).
\end{equation}
The training loss is modified from \cref{eq:ddpm_loss}:
\begin{equation}
\label{eq:ddpm_loss_cond}
    \mathcal{L} = \mathbb{E}_{t\sim\left[1, K\right], \mbold{A_t}^0, \epsilon^k} \left[ \| \epsilon^k - \mbold{\epsilon_\theta}\right(\mbold{A_t}^0 + \epsilon^k, \mbold{O_t}, k\left) \|^2 \right].
\end{equation}

In practice, we exclude observation features from the denoising process for better accommodation of real-time robot control, while the formulation remains the same.

\section{Further Explanation of ControlNet-style Fine-tuning}
\label{app:method:zeroinit}
A common misunderstanding with zero-initialized weights and biases is that they produce zero gradients and are, therefore, untrainable. We demonstrate that the additional KV-projection layers $\left(\mathbf{W_z}, \mathbf{B_z}\right)$ and the object-centric representations $\mbold{Z}$ can be optimized despite their zero initialization, which is similar to the case in ControlNet~\citep{zhang2023adding}. 

Let $\frac{\partial \mathcal{L}}{\partial \mathbf{V_z}}$ denote the upstream gradient from the loss $\mathcal{L}$. The gradients for $\mathbf{W_z}$ and $\mathbf{B_z}$ are: 
\begin{equation}
\left\{
\begin{aligned}
    \frac{\partial \mathcal{L}}{\partial \mathbf{W_z}} &= \sum_{p,i} \frac{\partial \mathcal{L}}{\partial {\mbold{V_z}}_{p,i}} \cdot {\mbold{Z}}_{p,i} \\
    \frac{\partial \mathcal{L}}{\partial \mathbf{B_z}} &= \sum_{p,i} \frac{\partial \mathcal{L}}{\partial {\mbold{V_z}}_{p,i}} \cdot 1
\end{aligned}
\right.
\end{equation}
Since $\mbold{Z}$ is non-zero, $\frac{\partial \mathcal{L}}{\partial \mathbf{W_z}} \neq \mathbf{0}$ if $\frac{\partial \mathcal{L}}{\partial \mbold{V_z}} \neq \mathbf{0}$. Similarly, $\frac{\partial \mathcal{L}}{\partial \mathbf{B_z}}$ accumulates non-zero gradients. After one gradient step:  
\begin{equation}
\left\{
\begin{aligned}
    \mathbf{W_z}^* = \mathbf{W_z} - \beta_l \cdot \frac{\partial \mathcal{L}}{\partial \mathbf{W_z}} \neq \mathbf{0} \\
    \quad \mathbf{B_z}^* = \mathbf{B_z} - \beta_l \cdot \frac{\partial \mathcal{L}}{\partial \mathbf{B_z}} \neq \mathbf{0}
\end{aligned}
\right.
\end{equation}
This ensures $\mbold{K_z}^*$ and $\mbold{V_z}^*$ become non-zero, allowing the dual-attention to incorporate $\mbold{Z}$.

Considering $\mbold{Z}$ is learnable, its gradient is:
\begin{equation}
    \frac{\partial \mathcal{L}}{\partial \mbold{Z}} = \mathbf{W_z}^T \cdot \frac{\partial \mathcal{L}}{\partial \mbold{V_z}}.
\end{equation}
Since $\mathbf{W_z}^* \neq \mathbf{0}$, $\mbold{Z}$ receives non-zero gradients and is updated accordingly. This aligns with the zero-convolution principle, where gradients persist despite zero-initialized parameters. We fine-tune the expert policy using the conditional denoising loss as defined in \cref{eq:ddpm_loss_cond}.

With the ControlNet-style fine-tuning, we efficiently integrate additional object-centric conditions into the pre-trained visuomotor policy. This approach ensures that when the KV-projection layers are zero-initialized in the dual cross-attention module, the deep neural features remain unaffected prior to any optimization. The capabilities, functionality, and output action quality of the pre-trained visuomotor modules are perfectly preserved, while further optimization becomes as efficient as standard fine-tuning. This allows \method to simultaneously leverage the advantages of large-scale pre-training and object-centric representations, accelerating real-world robot adoption by significantly reducing the data requirements for task deployment.

\section{Implementation Details of \method}
\label{app:method:detail}
In the main paper Sec. 4.1, we pre-train the policy $\mbold{\pi_g}$ on the full DROID dataset~\citep{khazatsky2024droid}, using the wrist camera image $\mbold{I_t}$, end-effector poses and gripper widths $\mbold{q_t}$, and episode language descriptions $\ell_t$. The observation and action horizons are set to $T_o=2$ and $T_a=16$. The pre-trained policy, implemented as a Diffusion Transformer~\citep{chi2023diffusion} with 29M parameters, uses a CLIP~\citep{radford2021learning} ViT-B/16 vision encoder and a Transformer text encoder. We pre-train $\mbold{\pi_g}$ with AdamW (learning rates: $1 \times 10^{-4}$ for denoising model, $3 \times 10^{-5}$ for vision; text encoder frozen) on 4 NVIDIA A800 GPUs for 3 days.
In Sec. 4.2, we extract object-centric representations from raw images. In Sec. 4.3, we fine-tune $\mbold{\pi_e}$ on evaluation tasks, adding $\sim$5M parameters. Fine-tuning uses the same settings as pre-training and runs on a single NVIDIA A800 GPU for 12 hours.

\section{Details of Experiments}
\label{app:detail_exp}
\subsection{Data Collection}
We collect a small set of demonstrations for each evaluation task.
For short-horizon tasks, we use UMI~\citep{chi2024universal}, an arm-agnostic data collection system with a hand-held gripper for efficient demonstration gathering. UMI features a wrist-mounted GoPro camera that captures RGB images and 6D end-effector pose trajectories using visual SLAM~\cite{campos2021orb} fused with onboard IMU data.
For long-horizon tasks, we use Meta Quest~\cite{metaquest} to teleoperate the AstriBot-S1~\cite{astribot_s1}, enabling immersive, low-latency 6DoF control via VR motion tracking. The operator's hand movements are mapped to the robot in real time, allowing intuitive and precise demonstrations. AstriBot-S1 is a more human-like robot with spherical joints at the shoulder and elbow, closely mimicking the range and fluidity of human articulation.
The number of demonstrations per evaluation task is detailed in the main paper Tab. 1.

\begin{figure}[!t]
    \centering
    \includegraphics[width=1.0\linewidth]{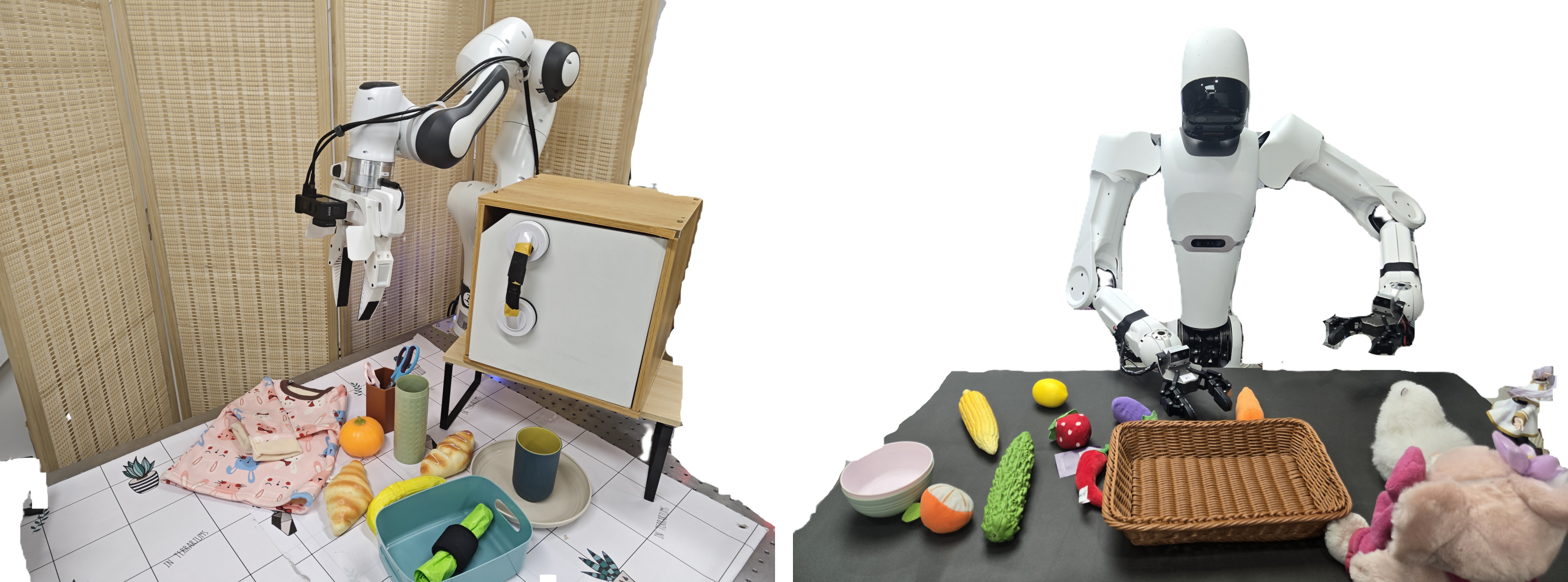}
    \caption{\textbf{Evaluation Setup.} The evaluation uses two robot platforms: the Franka Panda (left), for 6 short-horizon tasks; and the AstriBot-S1 (right), for 2 long-horizon tasks.}
    \label{fig:eval_setup}
\end{figure}

\subsection{Evaluation Setup and Protocol}
\begin{figure}[!h]
    \centering
    \includegraphics[width=\linewidth]{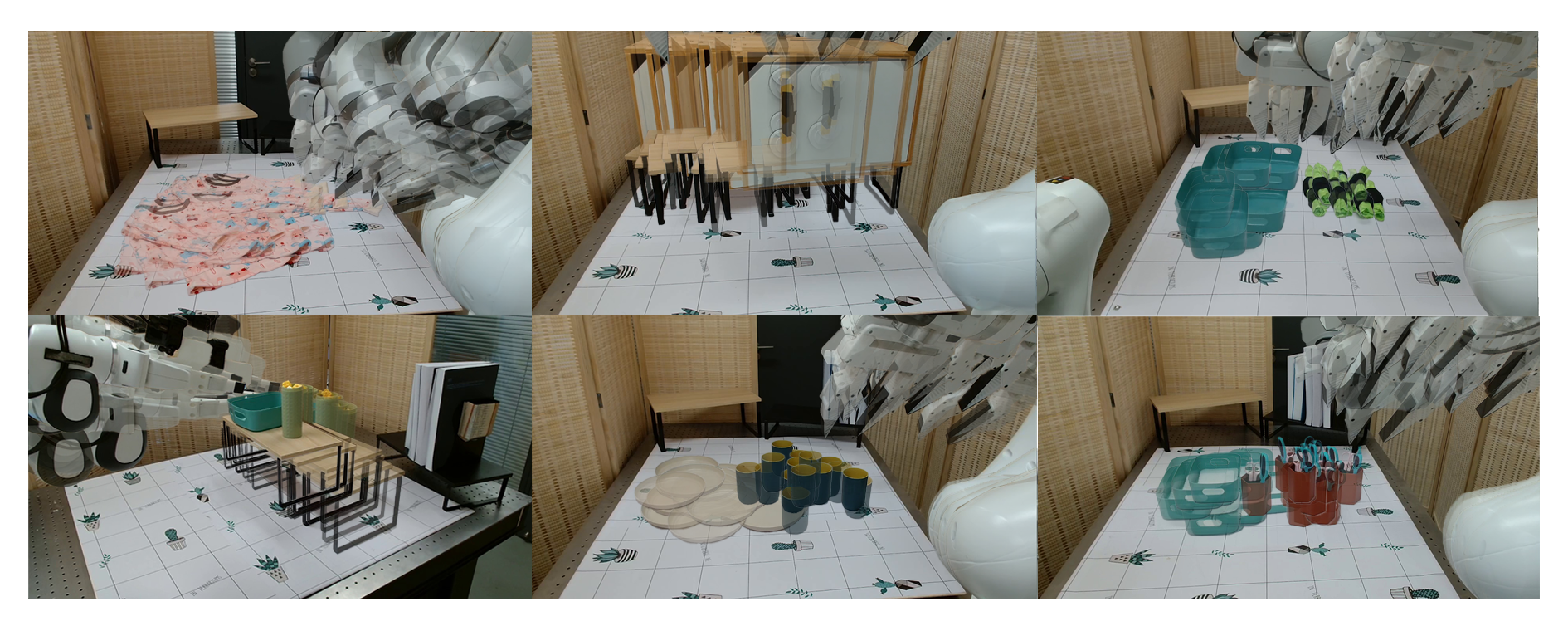}
    \caption{\textbf{Initial state distribution of policy evaluation.}}
    \label{fig:eval_initial}
\end{figure}

For short-horizon tasks, we deploy a Franka Emika FR3 arm with a Panda gripper and the same GoPro camera used during data collection for policy inference. For long-horizon tasks, we execute the policy on the AstriBot-S1 robot, which is equipped with a wrist-mounted RealSense camera for capturing RGB observations. Task success rate serves as the primary evaluation metric. Each trial is terminated if the policy shows no sign of progress, the robot enters a potentially unsafe interaction with the environment, or the task is completed. All evaluations are conducted in the same environment used for data collection, but with randomized initial configurations of both the robot and the objects to ensure robustness and generalization. \cref{fig:eval_setup} provides an overview of the evaluation setup, and \cref{fig:eval_initial} shows the initial objects and robot states distribution of policy evaluation.

\end{document}